\documentclass{article}
\usepackage{amssymb}
\usepackage{xcolor}
\usepackage{amsmath}
\usepackage{graphicx}
\usepackage{float}
\usepackage[stable]{footmisc}
\usepackage{algorithm}
\usepackage{algorithmic}
\usepackage[utf8]{inputenc}
\usepackage{caption}
\usepackage{subcaption}
\usepackage{hyperref}
\usepackage{comment}

\DeclareMathOperator*{\argmin}{arg\,min}
\graphicspath{figures_draft}

\title{A Genetic Algorithm approach to Asymmetrical Blotto Games with Heterogeneous Valuations}
\author{Aymeric Vi\'{e}}
\begin{document}
\maketitle

\noindent \textbf{Abstract} 
\noindent Blotto Games are a popular model of multi-dimensional strategic resource allocation. Two players allocate resources in different battlefields in an auction setting. While competition with equal budgets is well understood, little is known about strategic behavior under asymmetry of resources, as this situation is hardly solvable with analytical methods.  We introduce a genetic algorithm, a search heuristic inspired from biological evolution, interpreted as a mechanism of social learning, to solve this problem. Through multi-agent reinforcement learning, the algorithm evaluates the performance of random initial strategies. Most performant strategies are combined to create more performant strategies. Mutations allow the algorithm to efficiently scan the space of possible strategies, and consider a wide diversity of deviations. Iterating this process improves strategies until no profitable deviation exists. This method allows to identify optimal solutions to problems involving very large strategy spaces, such as Blotto games. We show that our genetic algorithm converges to the analytical Nash equilibrium of the symmetric Blotto game. We present the solution concept it provides for asymmetrical Blotto games. It notably sees the emergence of ``guerilla warfare'' strategies, consistent with empirical and experimental findings. The player with less resources learns to concentrate its resources to compensate for the asymmetry of competition. When players value battlefields heterogeneously, counter strategies and bidding focus is obtained in equilibrium. These features are consistent with empirical and experimental findings, and provide a learning foundation for their existence. These results open towards a more complete characterization of solutions in strategic resource allocation games. It places genetic algorithms as a search method of interest in game theoretical problems, to achieve further theoretical and operational achievements.\\



%\clearpage
%\tableofcontents
%\vspace*{\fill}

%\clearpage
%\listoffigures

\clearpage

\section{Introduction}

The Colonel Blotto game is an old game theory problem first introduced by \cite{borel}. It consists in a constant-sum allocation game in which two players, A and B, compete over $K$ independent battlefields. Each player has respectively $n_A$ and $n_B$ units of force to allocate simultaneously  across battlefields, without knowing the opponent's allocation strategy. Provided players respectively send $x_A$ and $x_B$ units on a given battlefield $k$, the player with this highest level of force wins the battlefield. Naturally, the when battlefields have the same value, the winner of the game is the player who conquers the highest number of battlefields.\\

Blotto Games constitute an archetype of a model of strategic resource allocation in multiple dimensions \cite{roberson}, with various empirical applications. Blotto Games are classically related to military conquest \cite{gross}. One can imagine a conflict in which two generals allocate armies across battlefields. Each one, or territory, has a value, corresponding for any characteristics the generals may find relevant: some may value territory facilities, others a strategic location. It is imaginable that these values could be private if the generals do not know each other; but also that they could be of common knowledge: general A values territory economic features, but knows general B values the location. Another frequent application of Blotto games is electoral competition in majoritarian systems \cite{myerson93, laslier2002}, in particular the US presidential election. Two players with specific, limited budgets compete over a number of states/constituencies, each rewarding the winner with a given number of Electoral College members. How these members are valued may differ: a party may find more interest and prestige to win in some state. Likewise, we can conceive situations in which this difference in valuation is private, as well as scenarii in which it falls under common knowledge. Finally, some particular auction settings fit nicely into this concept. For instance, auction over Sports TV bundles by main broadcasters often take the form of simultaneous auctions with several bundles being offered. Broadcasters with limited, possibly asymmetric budgets, may have specific valuations for the bundles. These specificities (desires of the particular group of customers,expected revenues from advertising...) may be private or public to the players.\\

In the traditional approach of Blotto games introduced since \cite{borel}, battlefields are assumed to have the same value, hence the payoff of the game is proportional to the fraction of the battlefields won \cite{roberson}. Winning the game is a byproduct of maximizing one's score, which is the number or proportion of battlefields won. In this Blotto Game with homogeneous battlefields values (also denoted weights in the literature), an equilibrium in this game consists in a pair of $K$-variate distributions. First solutions were provided by \cite{borel_ville} for a Blotto Game with 3 battlefields, and symmetric competition between players, i.e., $n_A$ = $n_B$. These solutions were extended by \cite{gross} for any $n \geq 3$, still under this symmetry setting. In this setting, \cite{roberson} more recently identified solutions for both symmetric and asymmetric Blotto Games under the assumption of identical battlefields weights. Roberson notably showed that uniform uni-variate marginal distributions were a necessary condition for equilibrium, an intuitive result as the battlefields' values are considered to be identical.\\

This paper departs from the assumption of identical battlefields values across players. We first consider that valuations of battlefields are not identical. Even in the most typical examples given to illustrate Blotto Games, battlefields values are heterogeneous: some battlefields are more valuable to one or the two players than others. The US Presidential election features 51 states, each bringing a different number of members of the Electoral College upon party victory in the state, a special feature fueling competition over the so-called ``swing states'' \cite{garand}. 
Second, we consider the possibility that battlefields valuations may vary across players. Player A and player B preferences as to what value they think each battlefield is worth have no reason to be identical. We can easily conceive in our example of the US Presidential election that Democrat and Republican parties have different interests in mind, some preferred states they would enjoy winning (or not losing) more than in others. We consider in this paper the problem of finding a solution, or at least, equilibrium concepts, to asymmetrical Blotto Games with heterogeneous valuations. Third, we will as well consider the situation of asymmetrical competition, that is, when players compete with different resource endowments, a game variant that is less understood in the literature. \\

\begin{figure}[H]
    \centering
    \includegraphics[scale = 0.55]{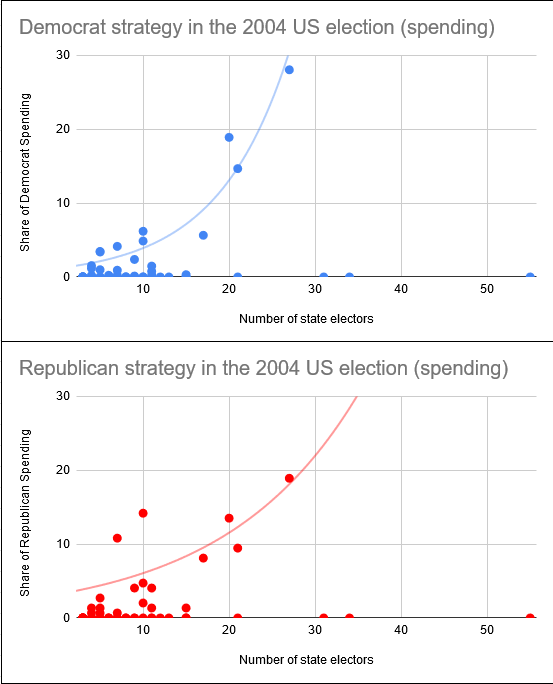}
    \caption{Party spending share in TV advertising per state vs state Electoral College endowment in the 2004 US Presidential Election. Data from \cite{USA_FEC}, with exponential fit.}
    \label{fig1}
\end{figure}

Empirical Blotto Games are characterized by special stylized facts, or behavioral puzzles. In experiments of asymmetrical Blotto games, disadvantaged players concentrates resources more than the advantaged player, developing some ``guerilla warfare'' strategies \cite{compte2019, chowdhury}. In settings prone to heterogeneity of battlefields valuations such as the US Presidential election, players do not spend on what they seemingly should value most \textit{in absolute terms}. In the 2004 election e.g., as illustrated in Figure \ref{fig1}, Republicans did not spend any dollar in TV advertising in Texas, neither did Democrats in California, even though these states respectively contribute by 38 and 55 Electoral College members. At the same time, both parties spent almost 30\% of their advertising budget in Florida, which ``only'' grants 29 members. A second objective of using our genetic algorithm process is to reproduce, and by so doing, understanding better, why these behaviors emerge, and how they can be rationalised in the context of a learning process.

Under these relaxed assumptions make identifying a solution, an equilibrium to these Blotto Games, and understanding these behaviors, a challenging task. Asymmetry in resources and in valuations make traditional analytical methods difficult to implement. In addition, they make the space of possible strategies very large, and richer. Competition does no longer only deal with competing over a given number of similar battlefields: it now involves identifying the important battlefields, and understanding what the other player's preferences are, in order to maximize one's payoff in the game and chances of victory. Recent progress from computer science has shown ability of alternative methods to thrive in environments of high strategical complexity such as go, chess, shogi and wargames \cite{silver1,silver2,silver3,vinyals}. To be able to characterize equilibrium distributions in the Blotto Game with heterogeneous valuations and asymmetric resources, we propose to rely on a related evolutionary process, inspired by the recent progress of combinations of algorithmic efficiency and games. This evolutionary process is consistent with viewing bounded rational players learning to find optimal strategies by strategy restrictions. The genetic algorithm does not consider all possible strategies before making its choice, but evaluates the performance of its initial candidate strategies, combine the best to form new ones, and locally perturbs these strategies to explore alternatives, analogously to a social learning environment. Some of the most popular evolutionary algorithms are genetic algorithms, introduced by \cite{holland}. As a search method inspired from Darwin's evolution theory, they reflect natural selection (fittest individuals are selected from reproduction and offspring production). They have previously been used with game theory as social learning mechanisms \cite{dawid}, converging to near Nash states as economic agents want to obtain the best strategy with respect to an objective function and the strategies of others \cite{riechmann}. Two important contributions of genetic algorithms (GA) to game theory include the study of auctions with artificial adaptive agents \cite{andreoni} and the evolution of strategies in iterated prisoner's dilemma \cite{axelrod}, showing the ability of learning by GAs to come to sophisticated, close to equilibrium, strategies and behaviors.\\

We show in this article that this approach can relevantly be undertaken to study Blotto games. Learning under a genetic algorithm provides a search heuristic that is showed capable to converge to the analytical Nash equilibrium, in the situations in which it is known. The genetic algorithm is successful in scanning a space of possible strategies of a considerable size and dimensionality in a finite time and with finite capacities, to converge independently to the Nash equilibrium of the game. This enforces the validity of the approach, in its attempt to tell us something about equilibrium and resource allocations in situations of asymmetry and heterogeneity of valuations, in which we do not have analytical solutions for the Nash equilibrium. Running the genetic algorithm in these games, we discover that it generates ``guerrilla warfare'': concentration of resources on highest-valued battlefield when running with a resource disadvantage. It also develops sophisticated counter strategies, and bidding behavior that is consistent with empirical stylized facts in real-world typically modeled with the Blotto games framework. By so doing, the genetic algorithm provides both a solution concept for asymmetrical Blotto games with heterogeneous valuations, and insights to understand empirical behavior puzzles. It opens genetic algorithm as an operational method to tackle empirical problems modeled as Blotto games.\\

% \textcolor{red}{ivethereaderapreviewofyourmain(theoretical) results. Briefly mention how these results affectthe existing understanding of your question or a broadersubject your questions fits into. Give the reader a roadmap forrest of the paper.} TAlgorithm strategies reproduce the phenomena of concentration of resources and guerilla warfare in situation of disadvantage. It also reproduces the apparent irrational bidding, which is rather explained in our setting by the players identifying battleifleds net valuations ...\\

This article is organized as follows. We first present our model of Blotto Game in section \ref{section_model}. The genetic algorithm is detailed in section \ref{section_algorithm}. Section \ref{section_results} presents the findings of the genetic algorithm. Notably, convergence to the analytical Nash equilibrium in analytically solvable Blotto game variants, and propositions of optimal strategies in asymmetric and heterogeneous cases, generating concentration of resources and sophisticated counter strategies in equilibrium. Section \ref{section_conclusion} concludes.

\section{Model}
\label{section_model}
\subsection{The general Blotto Game}

\textit{\textbf{Players}} Consider the following classical setting. Two players, denoted $A$ and $B$, are competing in an auction setting on several battlefields. With limited resources $n_i, i \in \{A,B\}$, they simultaneously allocate resources in each battlefield. Depending on the relative resource endowments of players, the Blotto game is called \textit{symmetric} if $n_A = n_B$, or \textit{asymmetric} when $n_A \neq n_B$.\\

\noindent\textbf{\textit{Valuations}} Each battlefield grants each player a specific payoff, that we denote \textit{valuation}. These valuations are usually considered \textit{homogeneous}: constant across battlefields and players in classical versions of the Blotto Game. Here, we suppose instead that valuations are \textit{heterogeneous}: they may differ depending on battlefields, and also across players. These valuations are characterized by density functions $g_1$ and $g_2$. Specifically in this work, valuations are assumed to be distributed uniformly in $[0,1]$, and independently across players. Denoting $v_h^i$ the valuation gained by player $i$ upon winning battlefield $h$ by an auction process, the vector $\textbf{V}_i$ of player $i$'s valuations over $k$ battlefields reads:

\begin{equation}
    {V}^i = (v_1^i, \dots, v_h^i, \dots, v_k^i), i \in \{A,B\}
\end{equation}

\noindent\textbf{\textit{Strategies}} Each \textit{strategy} $\sigma$ in the Blotto game is a vector of allocations of resources. Each element specifies a given quantity of resources to be allocated on a given battlefield. A strategy $\sigma^i$ of player $i$ in a Blotto game with $k$ battlefields and resources $n_i$ takes the following form:

\begin{equation}
   {\sigma}^i = (x_1^i, \dots, x_h^i, \dots, x_k^i), i \in \{A,B\}
\end{equation}

In incomplete information Blotto games, the strategy can be represented as a function of $v$. Strategies ${\sigma}^i$ are subject to a resource constraint: players have limited resources, and cannot allocate more than their endowment. We make here a classical assumption in the Blotto game literature to not attribute any value to unspent resources: resources are here encouraged to be fully deployed, and players do not gain any utility by saving these resources \cite{roberson}. Formally, in a discrete competition case, with our expression of strategies, we impose that for each strategy $\sigma^i$:

\begin{equation}
\label{resource_constraint_discrete}
    \sum_{h=1}^k x_h^i \leq n_i, \textnormal{ all } i 
\end{equation}
    
Considering a continuum of battlefields in $(0,1)$, we define $\sigma^i(v_i)$ the strategy of player $i$ taking player own valuation $v_i$ as argument\footnote{Under complete information, the strategy can be represented as a function of the valuation vector of both players.}. in this context, a strategy is no longer an allocation vector, but a continuous function $\sigma^i: v \in (0,1) \xrightarrow{} \sigma^i(v_i) \in [0,n_i]$. The resource constraint under the continuum assumption can be written as follows\footnote{Recall that $g_i$ are the density functions of the distributions of the valuations of player $i$.}.

\begin{equation}
\label{resource_constraint_continuum}
    \int_{v} \sigma^i(v_i)g_i(v_i)dv_i \leq n_i \textnormal{, all $i$}
\end{equation}

\noindent\textbf{\textit{Auction contest function}} We assume, as in the classical Blotto game setting, that battlefields are awarded to the player with the highest allocation of resources\footnote{A known recent variant of this game introduces a lottery setting, in which the probability of winning a battlefield is a continuous function taking player's and opponent's allocations as arguments \cite{osorio, xu}}. In a wargame analogy, general A would win territory $k$ by deploying at least one more solider than general B. In the case of a tie, the battlefield valuation is attributed to a random player. Formally, we write the expected utility of battlefield $h$ of player $i$ with strategy $\sigma^i$ against player $j$ with strategy $\sigma^j$:

\begin{equation}
\label{utility}
    E(U^i_h(\sigma^i,\sigma^j)) = 
    \begin{cases}
    & v^i_h \textnormal{ if } x_h^i > x_h^j \\
    & \frac{1}{2}v^i_h \textnormal{ if } x_h^i = x_h^j \\
    & 0 \textnormal{ if } x_h^i < x_h^j \\
    \end{cases}
\end{equation}

\noindent\textbf{\textit{Maximization problem}} In equation \ref{utility}, winning battlefield $h$ grants the successful player $i$ a utilityof $v^i_h$, its valuation for battlefield $h$. Valuations of won battlefields add up to form a score, which is the object players are maximizing. Players aim at maximizing the sum of such expected utility, over the full range of the $k$ battlefields, conditional on their valuation vector $V^i, i \in \{A,B\}$. In the discrete battlefields case, the maximization problem of player $i$ can be formulated in terms of expected utility:

\begin{equation}
    \max_{\sigma^i} \sum_{h=1}^k E(U^i_h(\sigma^i,\sigma^j)) \textnormal{ under (\ref{resource_constraint_discrete})}
\end{equation}

Over a continuum of battlefields, and under mutual perfect information of homogeneous\footnote{Perfectly correlated across players}valuations, this maximization problem takes the form: %, for each $v^i$:

\begin{equation}
    \max_{\sigma^i(v^i,v^j)} \int v^ig_i(v^i)P\left(\sigma^i(v^i) > \sigma^j(v^j)\right) dv^i \textnormal{ under (\ref{resource_constraint_continuum})}
\end{equation}

\subsection{Information structure}

\noindent\textbf{\textit{Perfect information}} constitutes a benchmark of interest. We here consider that each player initially knows its valuation vector, and not the one of the other player. However, with experience, and by a revealed preference argument, these adverse valuations become known. Observing how much the opponent deploys for some battlefields, and playing against the resulting strategies, gives our learning player more and more precise guesses on adverse valuations, realizing a situation of perfect information on valuations.

% This assumption of private information is orthogonal to the assumption of homogeneous or heterogeneous valuations. This may be the most frequent setting in real world Blotto games: often, players know with good certainty what each battlefield, item, object is worth to them. Having preferences that satisfy some regularity conditions appears essential to strategic decision making. There are many reasons to believe that in spite of this private information, there may be uncertainty on the valuations of the other players, partial or complete. In a US Presidential election setting, both players know how many Electoral College members the other party could give by winning a specific state. They do not know however, some psychological factors, that may impact the preferences of the other party to spend on campaign expenses in a specific state. Perhaps it is very important for a given candidate to win in her native state, or instead to create voter turnout in another candidate's constituency. These phenomena make the private information case worth of study.

It also models a situation with empirical relevance. Consider again the example of the US Presidential election. This would be a good theoretical example of electoral competition between rational players with perfect knowledge of the game environment. A variable of interest to players, would be the number of Electoral College members each state attributes upon victory in the state. Such information is easily accessible, and both players certainly know with certainty these numbers, and know that the other party knows these numbers too. But through repeated plays of this game, less accessible elements of individual players' preferences would become known. Over the last few elections, Republicans would have understood Democrats' priority targets, spending schemes, tactics, psychological interests and conversely. Perhaps a given state, a given subset of the population, constitutes to either player an element of valuations, and this element will be gradually internalized by the opponent, as learning progresses. Perfect information also allows max-min strategies \cite{ahmad}, in which agents use their knowledge on valuations to prevent their opponent from winning a battlefield the opponent highly values. In our learning process, we obtain the perfect information benchmark by setting players' valuations in the beginning of the process to be stable over time, without drawing them again. This ensures that with time, agents' mutual preferences become known, and realise this perfect information situation.

\subsection{Symmetric equilibrium in a continuum of battlefields\footnote{With great courtesy of O. Compte.}}

Recall that $\sigma^A$ is the strategy of player A, and $\sigma^B$ the strategy of player B. In the continuum case, these strategies are a function taking as argument a battlefield valuation, and returning a resource quantity to allocate. We can reasonably assume that they are strictly increasing, and that the valuations of both players follow the same distribution, that we assume to be uniform over $[0,1]$, with density $g$ and cumulative density $G$. \\

Let $h(v) = (\sigma^B)^{-1}$. When player A bids a quantity $x$ on a battlefield, the probability of A winning the battlefield can be expressed as $x > \sigma^B(v)) = G(h(x))$. We can write player A's utility as:

\begin{equation}
    U(\sigma^A,\sigma^B) = \int_{v^A} v^AG(h(\sigma^A(v^A)))g(v^A)dv^A \textnormal{ under } (\ref{resource_constraint_continuum})
\end{equation}

Which constitutes a sum over all realizations of $v^A$. The first order condition for a particular $v^A$ reads, denoting $\lambda$ the Lagrangian multiplier associated with the constraint \ref{resource_constraint_continuum}:

\begin{equation}
\label{foc_continuum}
    v^Ah'(\sigma^A(v^A))g(h(\sigma^A(v^A))) = \lambda
\end{equation}

Under the assumption of a symmetric Blotto game with homogeneous valuations, resources are identical and the players face the same maximization problem, under identical constraints. As a result, the equilibrium will be symmetric: $\sigma^A(v) = \sigma^B(v)$, all $v$. As a result, $h(\sigma^A(v^A)) = v^A$ and (\ref{foc_continuum}) gives in equilibrium:

\begin{align}
    (\ref{foc_continuum}) \iff v^A \frac{1}{\sigma^{'A}(v^A)}g(v^A) = \lambda
\end{align}

Which gives us over all realizations of $v^A$:

\begin{equation}
\label{general_continuum_solution}
    \sigma^A(v) = n_A \frac{\phi(v)}{\int \phi(v)g(v)dv} \textnormal{ with } \phi(x) = \int_0^{x}vg(v)dv
\end{equation}

Let us now simplify this expression under the assumption that valuations are distributed according to a uniform (0,1) law. As a result, the general solution (\ref{general_continuum_solution}) translates into:

\begin{equation}
\label{continuum_solution_uniform}
    \sigma^A(v) = n_A \frac{v^2}{\int v^2dv}
\end{equation}

\begin{figure}[H]
    \centering
    \includegraphics[scale = 0.75]{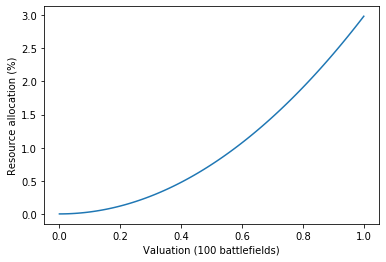}
    \caption{Symmetric equilibrium allocation (in share of total resources) over a quasi continuum of battlefields, $n = 100$}
    \label{fig2}
\end{figure}

We can deduce from (\ref{continuum_solution_uniform}) that equilibrium strategies in the symmetric Blotto game over a continuum of battlefields consists in allocating for any battlefield $h$ a fraction of the total resources equal to the ratio of the square valuation of $h$ by the sum of square valuations of all battlefields. This can be interpreted as a ratio of relative importance, where the nonlinear term has the players devote significantly more resources to more valuable battlefields. If we implement this strategy over 100 battlefields, the resulting allocation takes the form showed in figure \ref{fig2}, characterized by a strong, nonlinear concentration of resources on highest-valued battlefields. Let us outline that in this equilibrium, the top 50\% of battlefields by valuations receive more than 84\% of total resources, and that the 10 highest-valued receive alone almost 25\% of resources. The resulting pattern closely resembles empirical strategies deployed in real world strategic situations, such as the US Presidential election illustrated in Figure \ref{fig1}, with minor differences that we will explain further below. 

%%%
\subsection{A 3-battlefield simple case study}
\label{subsection_3battlefield}

\noindent\textit{\textbf{Model}} In the continuity of the general Blotto game introduced above, let us focus our attention on a simpler environment to allow us to better grasp the intuitions of the game and of its equilibrium. Let us consider that the two players indexed by $i \in \{A,B\}$, only compete over 3 battlefields. Suppose that the competition is symmetric: $n_A = n_B = n$. Suppose that we modify the utility function to include a more competitive setting: agents only aim at winning, i.e. having won more valuations than their opponents. This connects this model with the standard Blotto game with homogeneous weights on battlefields, in which score maximization and win are equivalent. 

\begin{equation}
\label{utility_modified}
    U^i(\sigma^i,\sigma^j) = 
    \begin{cases}
    & 1 \textnormal{ if } \sum_h U_h^i(\sigma^i,\sigma^j) > \sum_h U_h^{j}(\sigma^i,\sigma^j) \\
    & \frac{1}{2 } \textnormal{ if } \sum_h U_h^i(\sigma^i,\sigma^j) = \sum_h U_h^{j}(\sigma^i,\sigma^j) \\
    & 0 \textnormal{ if } \sum_h U_h^i(\sigma^i,\sigma^j) < \sum_h U_h^{j}(\sigma^i,\sigma^j) \\
    \end{cases}
\end{equation}

\noindent\textbf{\textit{Symmetric equilibrium in complete information}} Assume that $V^A = \left(0, \frac{1}{3}, \frac{2}{3}\right)$ and $V^B = \left(\frac{1}{3}, 0, \frac{2}{3}\right)$. Further suppose that these valuations are known by the players. The allocation $\sigma^* = (0,0,n)$ is a symmetric pure strategy Nash equilibrium. This symmetric equilibrium indeed grants in expectation a profit of $\frac{1}{2}$, and no deviation can improve this expected payoff. \\

%\noindent\textbf{\textit{Symmetric equilibrium in incomplete information}} Assume that $V^A = \left(0, \frac{1}{3}, \frac{2}{3}\right)$ and that $V^B$ is uncertain to $A$, and follows a uniform $(0,1)$ distribution. An equilibrium strategy for $A$ will put zero weight on the first battlefield with valuation 0. Resources of $A$ will be uniformly split on the two other battlefields: to guarantee in expectation the highest profit, bidding $nv^A_h$ on each battlefield $h$ maximizes expected profit over possible realisations of $V^B$, and likewise symmetrically for player $B$.\\ 

% \noindent\textbf{\textit{Symmetric equilibrium in complete uninformation}} When both $V^A$ and $V^B$ are unknown to the players, an optimal allocation over possible realisations of ($V^A$,$V^B$) mimics the distribution of valuations. Strategies would be uniform, and each battlefield allocation would be equal to $\frac{n}{k}$. This allows the best expected profit over all possible values taken by the valuations: the symmetric equilibrium in complete uninformation guarantees a utility value of $\frac{1}{2}$. Per instance, a possible deviation to playing $(0,0,n)$ (focusing on winning with certainty a specific battlefield, hoping that it is one with high value) only grants $\frac{4}{9}$ in expectation. \\

\subsection{Asymmetrical Blotto games and evolutionary equilibrium}

\noindent\textbf{\textit{Impossibility of analytical determination}} For both the general case, and the 3-battlefield simplified approach of Blotto games, analytical equilibria can be analytically identified. As both players have the same resources and distribution of valuations, they are indeed solving the same problem. This identity unravels as we relax the assumption of symmetric resources, to study asymmetric Blotto games. In the continuum case, the ability to pin down an equilibrium bidding strategy relies on the ability to simplify the first order condition \ref{foc_continuum} by symmetry of bidding functions in equilibrium. When resources are asymmetric, this simplification is no longer possible. Analytical determination of equilibrium without further constraints on the form of strategies then appears impossible.\\

Identifying equilibrium strategies in asymmetrical Blotto games characterized by a very large strategy space, can be achieved by the use of learning. Let us frame our search for an equilibrium characterization of such an equilibrium as an evolutionary game, in which equilibrium is obtained by learning. \\

\noindent\textit{\textbf{An evolutionary game}} Suppose our players $A$ and $B$ are repeatedly playing Blotto games against each other. Time, also denoted iterations and indexed by $t$, is discrete, and at each period, a Blotto game is played by the two players with \textit{families of strategies} $\Sigma^i_t, i \in \{A,B\}, t \in [0, 1, \dots, T]$ at a given time period, also denoted \textit{generation}. Families of strategies are simply vectors of strategies: a family $\Sigma$ contains a given number of strategies $\sigma$. Assuming that each family of strategies contains $p$ individual allocation strategies, we can formally note in the discrete battlefields case\footnote{In the continuum case, the evolutionary game formalisation of families of strategies can simply use the first line, in which families of strategies are vectors of $p$ different bidding functions.}:

\begin{align}
\begin{split}
  \left(\Sigma^A_t,\Sigma^B_t\right)  & = \left(\begin{bmatrix} \sigma^A_{t,1} \\ \sigma^A_{t,2} \\ \dots \\ \sigma^A_{t,p} \end{bmatrix}, \begin{bmatrix} \sigma^B_{t,1} \\ \sigma^B_{t,2} \\ \dots \\ \sigma^B_{t,p} \end{bmatrix} \right) \\
  & = \left(\begin{bmatrix} (x_{11}^A, \dots, x_{1h}^A, \dots, x_{1k}^A) \\ (x_{21}^A, \dots, x_{2h}^A, \dots, x_{2k}^A) \\ \dots \\ (x_{p1}^A, \dots, x_{ph}^A, \dots, x_{pk}^A) \end{bmatrix}, \begin{bmatrix} (x_{11}^B, \dots, x_{1h}^B, \dots, x_{1k}^B) \\ (x_{21}^B, \dots, x_{2h}^B, \dots, x_{2k}^B) \\ \dots \\ (x_{p1}^B, \dots, x_{ph}^B, \dots, x_{pk}^B) \end{bmatrix} \right)
\end{split}
\end{align}

Note that each allocation strategy $\sigma^i_{t,l}$ for all $i$, for all $t$ and for all $l \in \{1, \dots, p\}$ remains subject to the resource constraint (\ref{resource_constraint_discrete}). The evolutionary game is initialized with some starting families of strategies $(\Sigma^A_0,\Sigma^B_0)$.\\

Players are learning over time, in a related fashion to social learning: they play and experiment strategies against each other, measure the utility obtained from these allocations, and use this experience to improve their strategy for the next period. We denote this learning process $\mathcal{G}$. The evolution at time $t+1$ of a family of strategies $\Sigma^i_t$ is its transformation by the process $\mathcal{G}$. The joint evolution of the two players' families of strategies ($\Sigma^A_t,\Sigma^B_t$) can be described:
\begin{equation}
    (\Sigma^A_t,\Sigma^B_t) \xrightarrow{\mathcal{G}} (\Sigma^A_{t+1},\Sigma^B_{t+1})
\end{equation}

We propose in this work to implement the learning process $\mathcal{G}$ in the form of an evolutionary algorithm, a meta-heuristic search method inspired from biological evolution. In particular, we will implement this learning process in the form of a popular evolutionary algorithm: the genetic algorithm. This choice satisfies several requirements for our problem. It does not require any assumption on the underlying form of strategies. It is an adequate tool to solve optimization problems involving a large number of variables or constraints. Performing well on large space of possible solutions, genetic algorithms (GA) do not check all possible solutions, which for our problem would be computationally unfeasible, but permit to identify areas of interest for the solution. By so doing, the search for optimal solutions is computationally more efficient than classical exhaustive search methods, and exhibits a satisfying proximity to the concept of Nash equilibrium. Here, this learning specification allows players to experiment a diverse set of strategies contained in their family of strategies, and optimise the fitness of each generation, converging to the equilibrium of the asymmetrical Blotto game with heterogeneous valuations.

\begin{comment}
\textcolor{red}{need some refs here, but not too much, we have the introduction}
\end{comment}

\section{The Genetic Algorithm}
\label{section_algorithm}

We implement the learning process $\mathcal{G}$ as follows. First, we define some initial families of strategies, that are matrices of random strategies. Then, the following process is repeated. The fitness, that is, the utility, of each strategy of the family of strategy, is computed against every opponent strategy in her family of strategies. Best strategies are more likely to be selected as parents to generate child strategies, that save some of their allocations, and perform better in the Blotto game. Mutations allow to explore the diversity of the space of possible strategies, and to explore profitable deviations. We describe here in detail the functioning of the genetic algorithm.

\subsection{Initialization}

We start the learning process by defining initial families of strategies ($\Sigma^A_0,\Sigma^B_0$), matrices of $p$ strategies for both players. Recall that each row of the families of strategies is subject to the resource constraint (\ref{resource_constraint_discrete}). The family (or generations) matrices take the form:

\begin{equation*}
      \left(\Sigma^A_t,\Sigma^B_t\right)   = \left(\begin{bmatrix} (x_{11}^A, \dots, x_{1h}^A, \dots, x_{1k}^A) \\ (x_{21}^A, \dots, x_{2h}^A, \dots, x_{2k}^A) \\ \dots \\ (x_{p1}^A, \dots, x_{ph}^A, \dots, x_{pk}^A) \end{bmatrix}, \begin{bmatrix} (x_{11}^B, \dots, x_{1h}^B, \dots, x_{1k}^B) \\ (x_{21}^B, \dots, x_{2h}^B, \dots, x_{2k}^B) \\ \dots \\ (x_{p1}^B, \dots, x_{ph}^B, \dots, x_{pk}^B) \end{bmatrix} \right)
\end{equation*}

Each strategy (row) in the initial generation at $t = 0$ is determined as follows. We drop player superscripts for simplicity, assuming we describe here the determination of the initial family of $p$ strategies of a player with resources $n$, in a Blotto game with $k$ battlefields.

\begin{algorithm}
\caption{Generate a strategy ${\sigma}^i = (x_1, \dots, x_h, \dots, x_k)$}
\begin{algorithmic}
\REQUIRE $\sigma_i = (0, \dots, 0, \dots, 0)$
\WHILE{$\sum_hx_h \leq n$}
\STATE $a \in [1, k], a \in \mathbb{N}, a \sim U(1,k)$
%\STATE $b \in [0, n - \sum_hx_h], b \in \mathbb{N}$
%\STATE $x_a = b$
\STATE $x_a = x_a + 1$
\ENDWHILE
\RETURN ${\sigma}^i = (x_1, \dots, x_h, \dots, x_k)$
\end{algorithmic}
\end{algorithm}

\begin{figure}[H]
    \centering
    \includegraphics[scale = 0.6]{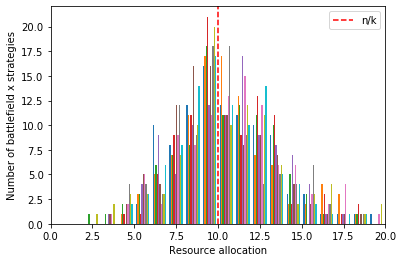}
    \caption{Distribution of resource allocations across battlefields in $\Sigma_0$}
    \label{fig3}
\end{figure}

This process is repeated $p$ times for each player to generate the first generation, i.e. the first family $\Sigma^i_t$. As a result, as Figure \ref{fig3} shows, all resource allocations points among the $pk$ battlefield-strategies appear initially normally distributed with mean $\frac{n}{k}$. 

\subsection{Strategy fitness}

Once this initial ecology of possible strategies is defined, we analyze the fitness of each one. That is, we analyze how well each strategy would perform if it was implemented in the Blotto game considered, playing against the strategies of the other player. This criterion is simply the utility as defined in (\ref{utility_modified}). Recall that:

\begin{equation*}
    \left(\Sigma^A_t,\Sigma^B_t\right) = \left(\begin{bmatrix} \sigma^A_{t,1} \\ \sigma^A_{t,2} \\ \dots \\ \sigma^A_{t,p} \end{bmatrix}, \begin{bmatrix} \sigma^B_{t,1} \\ \sigma^B_{t,2} \\ \dots \\ \sigma^B_{t,p} \end{bmatrix} \right)
\end{equation*}

Let us define the \textit{fitness vector} of each player $i$, at each period $t$, denoted $F^i_t$. It associates a given value to each strategy $\sigma^i_t$ of the generation $\Sigma^i_t$.

\begin{equation}
   F^i_t = \begin{bmatrix} U(\sigma^i_{t,1},\sigma^{-i}_{t}) \\ U(\sigma^i_{t,2},\sigma^{-i}_{t}) \\ \dots \\ U(\sigma^i_{t,p},\sigma^{-i}_{t}) \end{bmatrix}
\end{equation}

$U(\sigma^i_{t,l},\sigma^{-i}_{t}), l \in [1, \dots,p]$ is the utility for player $i$ from playing strategy $\sigma^i_{t,l}$ against all possible strategies of the other player at time $t$ as defined in (\ref{utility_modified}). This corresponds to the sum of utilities player $i$ would receive by playing a single strategy $\sigma^i_{t,l}$ against all strategies in the current generation of the other player: $\sigma^{-i}_{t,1}, \sigma^{-i}_{t,2}, \dots$, up to $\sigma^{-i}_{t,p}$. Formally:

\begin{equation}
   U(\sigma^i_{t,l},\sigma^{-i}_{t}) = \sum_{m=1}^pU(\sigma^i_{t,l},\sigma^{-i}_{t,m}) 
\end{equation}

As a result, the fitness vector $F^i_t$ allows to identify weakly dominant strategies in the family of strategies $\Sigma_t^i$. It provides a criterion to select the \textit{best} strategies at this period $t$, and by so doing, generate a generation $t+1$ with more optimal strategies. We introduce an \textit{elitist} component in this construction of the next generation: $\Sigma_{t+1}^i$ will contain without alteration the strategy with the highest fitness. It will also contain a mutated version of this current-best strategy. This avoids losing the current best solution from one period to the next, and permits even greater improvement of this current best result.

\subsection{Strategy crossover and mutation}

There exist several methods to select parents strategies in a genetic algorithm, responding to a necessary trade-off between the diversity of the solutions, and their fitness. The objective is to improve the performance of the generation of strategies, while maintaining enough ``genetic'' diversity to explore the space of possible actions. We adopt in this model a classical \textit{tournament method}. An arbitrary number of $3$ strategies is selected in generation $\Sigma_t^i$, and their fitness stored in $F^i_t$ is compared. The resulting two best are selected to be the \textit{parents} of a new \textit{child} strategy, denoted $\sigma_{t+1}^{'i}$. \\

How do we select parents? We choose here to adopt a fitness-based probabilistic selection. The higher the fitness of a strategy, the higher are the chances of it being selected to be the parent of new strategies. This allows to converge in a faster way than purely random selection of parent strategies. Formally, the probability of a strategy $\sigma_{t,a}^i, a \in [1,p], a \in \mathbb{N}$ with fitness $U(\sigma_{t,a}^i,\sigma_t^{-i})$ of being selected is equal to the ratio of its fitness to the sum of all strategy fitness. This ratio $r_a$ can be formalised:

\begin{equation}
\label{fitness_probabilistic}
    r_a = \frac{U(\sigma_{t,a}^i,\sigma_t^{-i})}{\sum_{l=1}^pU(\sigma_{t,l}^i,\sigma_t^{-i})}
\end{equation}

We proceed by a variant of the \textit{uniform crossover} method: each element $h = 1, 2, \dots, k$ in the child strategy vector is a weighted average of elements $h$ of parents strategies, where the weight $\omega$ is random, and distributed according to a uniform $(0,1)$ law. By so doing, the uniform crossover leading to the child strategy respects the resource constraint described in (\ref{resource_constraint_discrete}). Formally, with parents $\sigma_{t,1} = (a_1, a_2, \dots, a_k)$ and $\sigma_{t,2} = (b_1, b_2, \dots, b_k)$, the resulting child strategy is:

\begin{equation}
   \sigma_{t+1}^{'i} = \left(\omega a_1 + (1-\omega)b_1, \omega a_2 + (1-\omega)b_2, \dots, \omega a_k + (1-\omega)b_k \right)
\end{equation}

In the context of our genetic algorithm, we introduce some possibility of \textit{mutation of strategies}, under the resource constraint. With probability $\mu_t$ dependent on time, the child strategy mutates. While a wide diversity of mutation processes has been proposed in the literature for various solution concepts with genetic algorithms\footnote{We discuss in the later sections the importance for future research of a more comprehensive study of the properties of different mutation processes in the context of genetic algorithms.}, we here choose the random variant. Namely, if the strategy $\sigma_{t+1}^{'i}$ mutates, it is replaced by a random strategy determined as in Algorithm 1. This mechanism has the desirable property in the context of our problem that it contributes to identify a well-performing shape of the best response allocation. It is complementary to the crossover mechanism, that serves to adapt allocation amounts rather than the slope of allocations. However, what happens when the algorithm has identified the correct shape of resource allocations, but finds no profitable improvement from the flip bit mutation? In order to avoid stagnation on sub optimal strategies, while keeping the resource constraint holding, we additionally implement in the mutation process some crossover noise. A random element in $\sigma_{t+1}^{'i}$ is increased by an amount $\varepsilon \in (0,1)$, and another random element is decreased by the same amount. This allows the program, when close to an optimal solution (so that a random mutation is very unlikely to yield better performance), to keep improving its strategy.\\
%remove langdon 2010

This mutation, and the related choice of the mutation probability $\mu$, allows the genetic algorithm to explore some diversity of solutions in the space of possible strategies, without falling into pure random search. We desire that the algorithm experiments in the early stages of learning to identify an optimal allocation strategy. At the same time, once this strategy has been identified, we aim at saving it rather than seeing it lost by further mutations. We want the genetic algorithm to efficiently scan a very large space of possible strategies, without it turning into a pure random search heuristic. A choice of $\mu = \frac{1}{p}$ would guarantee that in average, one strategy in the family of $p$ strategies mutates. This may be enough for simple environments with a small number of battlefields, but probably not enough diversity for problems involving a larger number of battlefields. We hence set the mutation probability to $\mu = \frac{k}{p}$ to give the program a higher mutation probability as we deal with more complex problems. In average, $k$ of the $p$ strategies in a given family of strategy (or generation) will mutate. \\

% The choice of a non-uniform mutation probability $\mu_t$ in equation (\ref{mutation_probability}) satisfies these targets. This specification makes sure that in average, $k$ elements in the family of $p$ strategies mutate. This allows the program learning in higher-battlefield games, more complex, to experiment relatively more. In addition, the incidence of mutations decreases over time\footnote{This mutation probability can be related to a general form $\frac{k}{p {t}^\gamma}$, where $\gamma$ is tuned to leave the program enough time to come to the optimal allocation before stabilising. In our context, simulations for instance suggest that while $\gamma = 1$ leaves too short a window for exploration, $\gamma = \frac{1}{3}$ over-mutates the family of strategy, generating losses of optimal allocations.}, allowing the program to stabilise on optimal allocations identified in the earlier stages. The choice of an inverse rather than exponential or logarithmic function allows to keep a positive probability of mutation even after a sizeable number of periods of time, smoothing the stabilising mechanism to avoid exogenously-imposed convergence.

%\begin{equation}
%    \label{mutation_probability}
%    \mu_t = \frac{k}{p \sqrt{t}}
%\end{equation}

%\begin{figure}[H]
%    \centering
%    \includegraphics[scale = 0.6]{figures_draft/mutation_proba.png}
%    \caption{Mutation probability over time}
%    \label{mutation_proba}
%\end{figure}

These two steps, crossover and mutations, are formalized below in the form of pseudo-code. Note that this Algorithm 2 is repeated $p-2$ times, as the next generation $\Sigma_{t+1}$ of $p$ elements is constituted of $p-2$ rows-strategies subject to crossover and mutation, and two rows corresponding to the best fitness solution of the current generation $\Sigma_{t}$  based on $F^i_t$, and a mutated version of this best solution. In this algorithm, 3 parents with indices $a,b,c$ are selected according to the relative fitness probabilistic process described in equation (\ref{fitness_probabilistic}).

Once these steps, initialization, fitness evaluation and finally crossover and mutation, are achieved, we simply move to the next iteration, from $t$ to $t+1$. We do not specify a termination condition in the learning process, as the objective -equilibrium- is unknown. We  set the maximum number of iterations $T$ to be large enough, and devote some attention to the analysis of the influence of the chosen value of $T$ over the results.

\begin{algorithm}
\caption{Generate a child strategy ${\hat{\sigma}}^i = (x_1, \dots, x_h, \dots, x_k)$}
\begin{algorithmic}
\REQUIRE $\tau = (a,b,c) \in [1,p], (a,b,c) \in \mathbb{N}$
\IF{$U(\sigma^i_{t,c},\sigma^{-i}_{t}) = \argmin \{U(\sigma^i_{t,l},\sigma^{-i}_{t}), l \in \tau\}$}
\REQUIRE $\sigma^i_{t,a} = (a_1, \dots, a_h, \dots, a_k)$
\REQUIRE $\sigma^i_{t,b} = (b_1, \dots, b_h, \dots, b_k)$
\REQUIRE $\omega \in (0,1), \omega \sim U(0,1)$
\STATE $\sigma^{'i} = \left(\omega a_1 + (1-\omega)b_1, \dots, \omega a_k + (1-\omega)b_k \right)$
\ENDIF
\REQUIRE $\delta \in (0,1), \delta \sim U(0,1)$
\IF{$\delta \leq \mu$}
\STATE \textit{Algorithm 1}
%\STATE $\hat{\sigma}_i = (0, \dots, 0, \dots, 0)$
%\WHILE{$\sum_hx_h \leq n$}
%\STATE $a \in [1, k], a \in \mathbb{N}, a \sim U(1,k)$
%\STATE $x_a = x_a + 1$
%\ENDWHILE
%\RETURN ${\hat{\sigma}}^i = (x_1, \dots, x_h, \dots, x_k)$

\ELSE \STATE $\hat{\sigma}^i = \sigma^{'i}$
\ENDIF 
\end{algorithmic}
\end{algorithm}

\section{Results}
\label{section_results}

This section presents our main results. We do not here present algorithm best strategies for all possible games or settings. We rather illustrate the contribution of this genetic algorithm by highlighting several properties of interest, under a certain random seed to allow comparison. \par
We first establish that our learning process finds similar results to the analytical equilibrium, in Blotto games with private information that are analytically solvable. Subsection \ref{results_sub1} demonstrates such convergence in a symmetric 3-battlefield game. Subsection \ref{results_sub2} shows a similar convergence to the continuum equilibrium in a symmetric 10-battlefield Blotto game. This gives us great confidence that our algorithm can learn something on situations that are not analytically solvable. Next, we analyse competition in Blotto games under heterogeneity of valuations and asymmetry of resources. We put an emphasis on subsection \ref{results_sub3} on the concentration of resources by the disadvantaged player. We finally investigate in subsection \ref{results_sub4} the emergence of counter strategies, and bidding strategies that to explain some stylized facts of interest on empirical Blotto games.

%\textbf{\textit{A word on the curse of dimensionality}} and why we use very low $n$ to help the program performance\\

\subsection{Convergence to the Nash equilibrium in symmetric 3-battlefield Blotto games}
\label{results_sub1}

Consider a Blotto game with 3 battlefields, and suppose that players' valuations are homogeneous and realized by:
\begin{equation*}
    V^A = V^B = \left(0,\frac{1}{3},\frac{2}{3}\right)
\end{equation*}

And that resources are symmetric: $n_A = n_B = n = 1$. We know from subsection \ref{subsection_3battlefield} that an analytical symmetric pure strategy Nash equilibrium exists at $(0,0,n)$ in complete information. We show below that our algorithm converges to this strategy, with some standard settings: the algorithm considers (only) $p = 50$ strategies per generation, and has a mutation probability of $\frac{k}{p} = 0.06$. 

% Results are not similar regardless of the payoff function the algorithm is trained with: \textcolor{red}{utility function for winning} or \textcolor{red}{utility function for score}. When trained to maximise its score without consideration of victory, the algorithm converges much more slowly than the algorithm trained to win the game by outscoring its opponent. At iteration 1,000, the former exhibits an average best strategy close to [0,2,8]\footnote{\textcolor{red}{oddly close to the continuum equilibrium applied for a k=3 case, this convergence to [0.6, 2, 7.4] appears robust across different iteration horizons under the mutation probability of $\frac{k}{p}$}} when the latter plays a stronger [0,0,10]. \\

\begin{figure}[H]
\centering
     \begin{subfigure}[b]{0.3\textwidth}
         \centering
         \includegraphics[width=\textwidth]{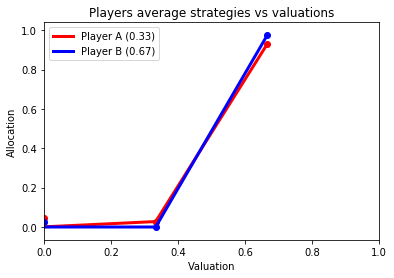}
         \caption{After 50 iterations}
         \label{fig5a}
     \end{subfigure}
     \hfill
     \begin{subfigure}[b]{0.3\textwidth}
         \centering
         \includegraphics[width=\textwidth]{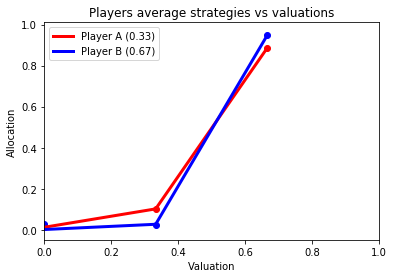}
         \caption{After 250 iterations}
         \label{fig5b}
     \end{subfigure}
     \hfill
     \begin{subfigure}[b]{0.3\textwidth}
         \centering
         \includegraphics[width=\textwidth]{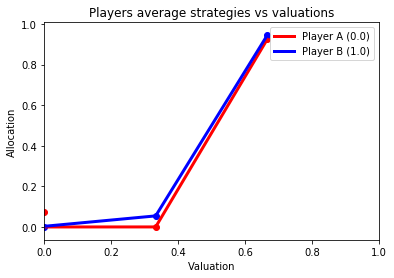}
         \caption{After 1,000 iterations}
         \label{fig5c}
     \end{subfigure}
        \caption{Convergence of average strategies in the symmetric 3-battlefield Blotto Game to the Nash equilibrium $(0,0,1)$}
        \label{fig4}
\end{figure}

As the valuations of both players become common knowledge, the algorithm converges very quickly to the identified pure strategy symmetric Nash equilibrium: $(0,0,n)$. In this simulation, players' best strategies in their family of strategies are $(0,0,1)$ as early as iteration 50. The average best strategy achieves over time this convergence towards $(0,0,1)$. It clearly shows evidence of a fast convergence to the Nash equilibrium, starting from random strategies at iteration 0. \\

What is the mechanism at play behind this convergence? Among these initial strategies, the ones putting the heaviest weight on the third battlefield valued $\frac{2}{3}$ by the two players directly obtain higher fitness, hence higher chances to be featured in the next generation of strategies. Mutations contribute to improve the strategy as soon as it reaches the symmetric Nash equilibrium $(0,0,1)$ at which no further improvement is possible. 

\subsection{Convergence to the continuum of battlefields equilibrium in symmetric Blotto games}
\label{results_sub2}

Let us run our learning process for a Blotto game with symmetric resources, homogeneous valuations across players, distributed uniformly in $(0,1)$. Without loss of generality, and in order to obtain ``well behaved'' valuation vectors that cover the full range of the uniform distribution interval for a low number of battlefields, we restrict our analysis to evenly spaced valuation vectors within the given interval. By so doing, we replicate the conditions of the continuum Blotto game, for which we identified the equilibrium in equation (\ref{continuum_solution_uniform}). Can our genetic algorithm converge near the symmetric equilibrium we identified?\\

Let us consider the following symmetric Blotto Game. The two players now compete over $k = 10$ battlefields, and have equal resources $n_A = n_B = n = 10$; family of strategies are composed of $50$ strategies, and valuations are considered homogeneous, that is, common among players. As usual, they are supposed to be distributed uniformly in $(0,1)$. The mutation rate is still set as $\frac{k}{p} = 0.2$.\\

After 1,000 iterations, results show that the genetic algorithm converges very closely to the Nash equilibrium, with pure learning, and absence of assumptions on the form of optimal strategies. That is, we have convergence without having the algorithm been ever exposed to the Nash equilibrium in any step of learning. The program develops a strategy that is very close to the analytical Nash for the continuum game, with slight variations that allows to actually beat this equilibrium strategy if it faced it in a Blotto game setting. We confirm here that our algorithm is capable to achieve without supervision or clues, the current best analytical solution of the symmetric Blotto Game. We are mostly looking at the average of \textit{average best} strategies to assess convergence and algorithm behavior. They indeed provide a more accurate and less variant outcome than temporary best strategies, or simply the average of strategies in a family of strategies. Both are indeed subject to temporary noise induced by particular mutations, or strategical artefacts. Average best strategies are the average of each best strategy at all iterations, and provide a more stable indicator of the outcome of the learning process. \\

\begin{figure}[H]
\centering
\includegraphics[width=.5\textwidth]{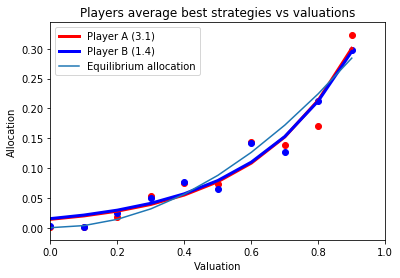}\hfill
\includegraphics[width=.5\textwidth]{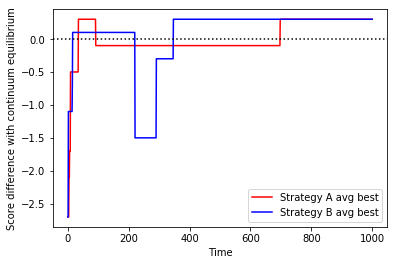}
\caption{Convergence of average best strategies to the continuum Nash equilibrium in a 10-battlefield Blotto Game with private information at 1,000 iterations}
\label{fig5}
\end{figure}

Right Figure \ref{fig5} plots the utility obtained by the algorithm strategy facing the equilibrium strategy, over time, showing progress through time and ability to perform eventually better than the current best analytical equilibrium. Figure panel \ref{fig5} also highlights some interesting features on the path to the convergence of the program. From initial random strategies, and only under the fitness and mutation process, the algorithm converges to a strategy that reassembles very closely to the Nash equilibrium over a continuum of battlefields, under symmetric competition and private information. Right Figure \ref{fig5} shows the relative performance of the algorithm against this continuum equilibrium. While in the first periods, the algorithm has a lower score (then loses the game) against this equilibrium, it gradually improves, with jumps likely corresponding to successful mutations, and manages to overcome the performance of the continuum equilibrium. The significant drop in performance around iteration 200 for the average best strategy B (in blue) probably corresponds to a local optimum, that the algorithm nevertheless manages to correct, as it improves its performance afterwards. This highlights a desirable property of the process: it appears capable of not falling into optimization traps and only local optima, and remains able and flexible enough to identify global optima with pure self play. 

\begin{figure}[H]
     \centering
     \begin{subfigure}[b]{0.3\textwidth}
         \centering
         \includegraphics[width=\textwidth]{figures_draft/Figure5a.png}
         \caption{$n = 1$}
         \label{fig6a}
     \end{subfigure}
     \hfill
     \begin{subfigure}[b]{0.3\textwidth}
         \centering
         \includegraphics[width=\textwidth]{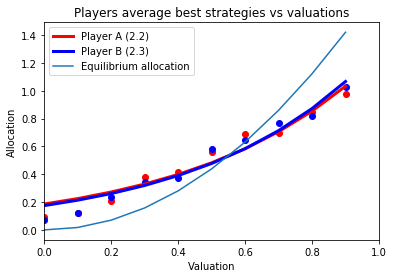}
         \caption{$n = 5$}
         \label{fig6b}
     \end{subfigure}
     \hfill
     \begin{subfigure}[b]{0.3\textwidth}
         \centering
         \includegraphics[width=\textwidth]{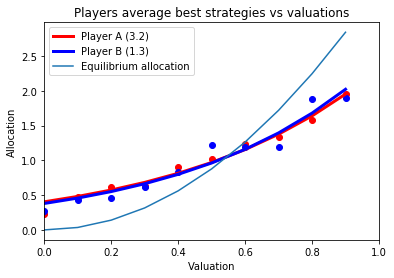}
         \caption{$n = 10$}
         \label{fig6c}
     \end{subfigure}
        \caption{Average best strategies at 1,000 iterations in the 10-battlefield game, with different values of $n$ illustrating the impact of strategy space dimensionality over convergence, every other parameter held constant.}
        \label{fig6}
\end{figure}

This 10-battlefield case, although not close to infinity, allows to observe the outcome of the genetic learning process over a much more complex setting than the 3-battlefield case above. It constitutes a less wrong approximation of the continuum case, than the previous 3-battlefield setting, and remains computationally reasonable for a finite time and a low number of strategies processed at any period of time. The dimensionality of the problem is indeed much higher as $k$ increases, and the space of possible strategies much larger. For this reason, reducing the dimensionality of the problem by adjusting the resource endowments $n$ of the players, without loss of generality, is crucial. This choice of $n$ has important consequences on the outcome of the genetic optimization process considered for at a given final iteration, and under a given common random seed\footnote{A random seed guarantees reproducibility of algorithm runs within an algorithm with some random components: number generation notably. A specific seed guarantees that the sequence of random number generated will be exactly the same, allowing comparison of different parameters setting with little impact of inconsistent random number sequence realisations.}. Consider the space of integer strategies under $k = 10$ battlefields and $n = 10$ resources available. This represents billions of strategies under the resource constraint, even restricting only to integer strategies. This allows to understand why for a larger number of battlefields such as $10$, dimensionality should be reduced as much as possible to help the algorithm handle the task in a short time. Figure panel \ref{fig6} shows how small differences in resources endowment $n$ impact the algorithm outcome, all other parameters held constant, as a result of the dimensionality of the problem.\\

% \textcolor{violet}{difference win/score figure panel. In 3b, there was a difference because of the discontinuity. With 10b, score and win give the same thing. SO we should totally forget about it and not write about it to not have people be confused?}

% \begin{figure}[H]
% \centering
% \includegraphics[width=.5\textwidth]{figures_draft/n1_score.png}\hfill
% \includegraphics[width=.5\textwidth]{figures_draft/n1_win.png}
% \caption{Strategies at 1,000 iterations based on score maximization (left) or win criterion (right), for 10 battlefields, $n = 1$. Both perform slightly better than the continuum equilibrium.}
% \label{fig12}
% \end{figure}

\begin{figure}[H]
     \centering
     \begin{subfigure}[b]{0.3\textwidth}
         \centering
         \includegraphics[width=\textwidth]{figures_draft/Figure5a.png}
         \caption{At 1,000 iterations}
         \label{fig7a}
     \end{subfigure}
     \hfill
     \begin{subfigure}[b]{0.3\textwidth}
         \centering
         \includegraphics[width=\textwidth]{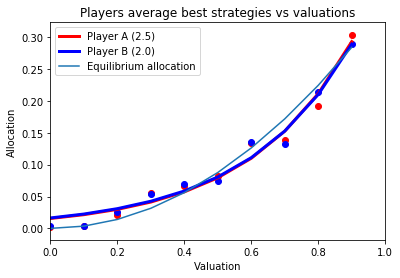}
         \caption{At 2,500 iterations}
         \label{fig7b}
     \end{subfigure}
     \hfill
     \begin{subfigure}[b]{0.3\textwidth}
         \centering
         \includegraphics[width=\textwidth]{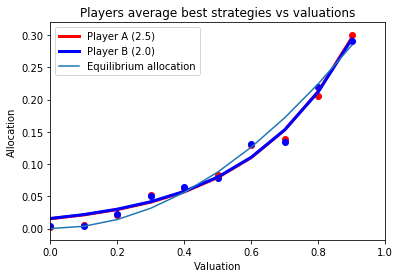}
         \caption{At 5,000 iterations}
         \label{fig7c}
     \end{subfigure}
        \caption{Average best strategies in the 10-battlefield game do not seem to change at different iterations horizons $T$}
        \label{fig7}
\end{figure}

\begin{figure}[H]
     \centering
     \begin{subfigure}[b]{0.3\textwidth}
         \centering
         \includegraphics[width=\textwidth]{figures_draft/Figure5b.png}
         \caption{At 1,000 iterations}
         \label{fig8a}
     \end{subfigure}
     \hfill
     \begin{subfigure}[b]{0.3\textwidth}
         \centering
         \includegraphics[width=\textwidth]{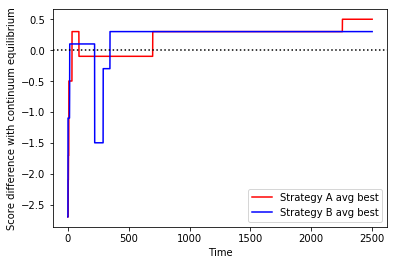}
         \caption{At 2,500 iterations}
         \label{fig8b}
     \end{subfigure}
     \hfill
     \begin{subfigure}[b]{0.3\textwidth}
         \centering
         \includegraphics[width=\textwidth]{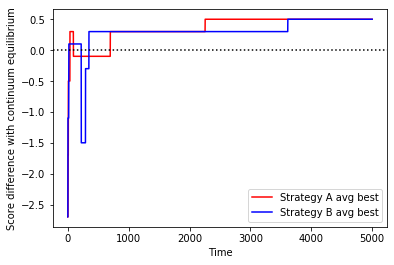}
         \caption{At 5,000 iterations}
         \label{fig8c}
     \end{subfigure}
        \caption{Difference in utility between average best strategy and the continuum equilibrium at different iterations horizons $T$}
        \label{fig8}
\end{figure}

To establish whether these strategy distributions are indeed convergent, we need to check what is the outcome of the learning process if we extend its number of iterations. Algorithm behavior does not appear to significantly change when we extend the iterations horizons. From Figure panel \ref{fig8}, we observe that running the program for a much higher number of periods seem to result in a minor improvement of the performance around iteration 2,250 (met by the other player later on around iteration 3,000), suggesting that the program found some more improvement and managed to probably win a more battlefield when confronted to the continuum equilibrium strategy. However, as the Figure panel \ref{fig7} shows, the converging strategy is not change. This allows us to consider with great confidence that the algorithm does converge, and that its outcome strategy is stable in regard of our learning process. Figure panels \ref{fig7} and below illustrate players' allocations strategies as a function of their own valuations: they do not directly indicate how resources allocations enter in conflict at the level of battlefields. We are here mostly interested in the characterization of the allocation strategy function as a function of own valuations, in our situation of imperfect competition.

%\textcolor{red}{what then? 10k? what is the end?}

\subsection{Concentration of resources in asymmetric Blotto Games}
\label{results_sub3}

Let us consider again a 10-battlefield Blotto Game with private information, and present the behavior of the genetic algorithm in the Blotto Game with asymmetric competition. The degree of asymmetry is characterized by the game parameter $\alpha$. We recall that $\alpha \in (0,1)$ is the fraction of the resources of player A that are left at B's disposition. When $\alpha = 1$, the competition is symmetric; for $\alpha = 0.5$, player B plays with half of player A's resources. We study how different levels of asymmetry in resources allow the algorithm to converge to different allocation strategies, notably under the lens of concentration of resources. The disadvantaged player may indeed attempt to concentrate resources on some specific battlefields. Does playing from behind indeed incentives players to concentrate more their allocations, and towards which battlefields? \\

We will present how in our genetic algorithm, this concentration of resources impacts our results when players have the same valuations in the next section \ref{asym_homog}, and when these valuations are allowed to be heterogeneous among them in the later section \ref{asym_heterog}.

\subsubsection{Concentration of resources under homogeneous valuations}
\label{asym_homog}

\begin{figure}[H]
     \centering
     \begin{subfigure}[b]{0.3\textwidth}
         \centering
         \includegraphics[width=\textwidth]{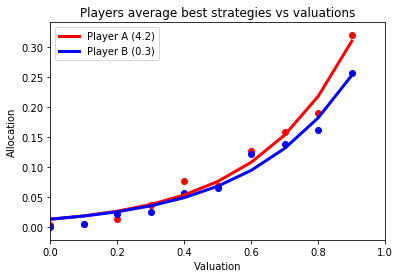}
         \caption{$\alpha = 0.85, \beta_A = 3.51, \beta_B = 3.27$}
         \label{fig9a}
     \end{subfigure}
     \hfill
     \begin{subfigure}[b]{0.3\textwidth}
         \centering
         \includegraphics[width=\textwidth]{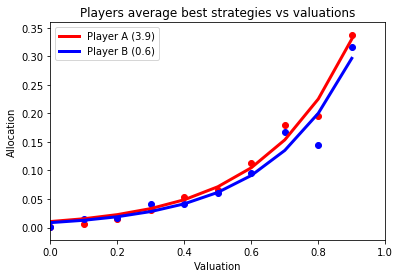}
         \caption{$\alpha = 0.9$, $\beta_A = 3.82, \beta_B = 3.92$}
         \label{fig9b}
     \end{subfigure}
     \hfill
     \begin{subfigure}[b]{0.3\textwidth}
         \centering
         \includegraphics[width=\textwidth]{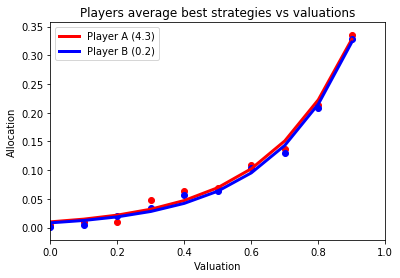}
         \caption{$\alpha = 0.95, , \beta_A  = 3.87, \beta_B = 4.06$}
         \label{fig9c}
     \end{subfigure}
        \caption{Average best strategies against valuations, with exponential fit, after 1,000 iterations}
        \label{fig9}
\end{figure}

When asymmetry in resources is rather low (for $\alpha = 0.9$ and $\alpha = 0.95$ in respective Figures \ref{fig9b} and \ref{fig9c}), $\beta_B > \beta_A$, showing that the disadvantaged player concentrates more its resources than the advantaged player. Throughout learning, low levels of asymmetry such as $\alpha = 0.95$ see the disadvantaged player win over the advantaged one. Indeed, its increased concentration of resources provides it with stronger strategies, the implementation of which is not too affected by the difference in resources as $\alpha$ is close to 1. We observe that low levels of asymmetry, can actually improve learning, and allow the disadvantaged player to ``learn faster'' than its advantaged opponent, as it faces a harsher environment.\\

For a greater level of asymmetry here illustrated by $\alpha = 0.85$ in Figure \ref{fig9a}, a different pattern emerges, and the disadvantaged player concentrates resources less than the advantaged player. We interpret this behavior as a consequence of a too large level of asymmetry. The difference in resources does not allow the disadvantaged player to still minimize its losses by focusing the highest-valued battlefield: its opponent has simply too much resources for this strategy to work. As a result, the only way for the disadvantaged player to still minimize its score loss, is to devote some more resources to the lowest-valued battlefields.

\subsubsection{Concentration of resources under heterogeneous valuations}
\label{asym_heterog}

Let us now turn our attention to a different setting, in which players' valuations of battlefields are allowed to be heterogeneous. 
Under our random seed, valuations vectors random realisations read:

\begin{equation*}
    V^A = (0, 0.1, 0.2, 0.3, 0.4, 0.5, 0.6, 0.7, 0.8, 0.9)
\end{equation*}

\begin{equation*}
    V^B = (0.7, 0.5, 0.4, 0.1, 0.2, 0, 0.8, 0.9, 0.6, 0.3)
\end{equation*}

Alternative seeds will generate a different set of valuation vectors. As we are here interested in evaluating how different levels of asymmetry in resources shape competition in Blotto Game, the following simulations will be executed all other parameters held constant, including this realisation of valuation vectors. The results nevertheless hold for any other seed. The particular one used here is only presented to allow reproductibility, and illustrate the results.\\
% Might want to reformulate this paragraph, seems unclear
Under heterogeneous valuations however, more strategies options become available to the players as the game environment complexifies. The players are no longer only identifying what battlefields are worth the most to both players; they are also learning how their opponent values them. For a given valuation, the degree of competition, i.e. the extent to which players will fight over this particular battlefield, will vary depending on their heterogeneity of valuations. A battlefield with a low value shared by the two players will appear harder to win that a battlefield that has some value for one, but no value for the other player. We have the intuition that in this situation, a rational player would make better use of its resources by shifting at some resources away from the risky, contested battlefields, towards the less risky, valuable battlefield. As this one exhibits a strong difference between valuations, it may not be contested by players. This is, in our view, the model interpretation as to why Democrats for instance would not try to compete in the Texan State, and Republicans in New York. 

\begin{figure}[H]
     \centering
     \begin{subfigure}[b]{0.3\textwidth}
         \centering
         \includegraphics[width=\textwidth]{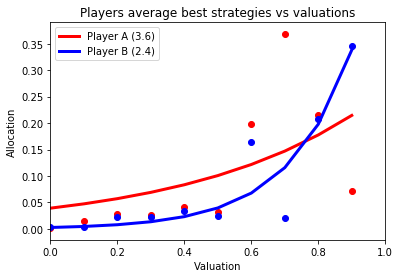}
         \caption{$\alpha = 0.85, \beta_B = 5.36$}
         \label{fig10a}
     \end{subfigure}
     \hfill
     \begin{subfigure}[b]{0.3\textwidth}
         \centering
         \includegraphics[width=\textwidth]{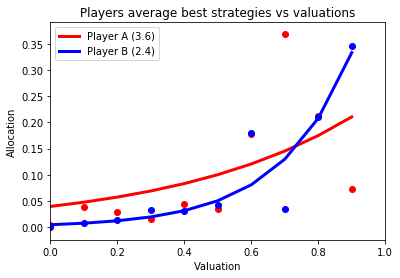}
         \caption{$\alpha = 0.9$, $\beta_B = 4.71$}
         \label{fig10b}
     \end{subfigure}
     \hfill
     \begin{subfigure}[b]{0.3\textwidth}
         \centering
         \includegraphics[width=\textwidth]{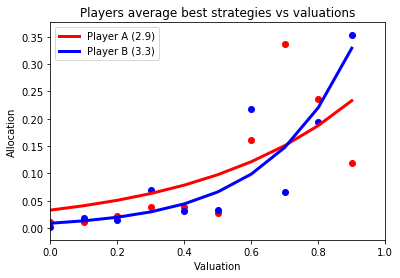}
         \caption{$\alpha = 1, \beta_B = 4.01$}
         \label{fig10c}
     \end{subfigure}
        \caption{Average best strategies against valuations after 1,000 iterations, with exponential fit}
        \label{fig10}
\end{figure}

Figure panel \ref{fig10} plots the strategies of the genetic algorithm in asymmetric competition with an exponential fit of the form $y = a e^{bv}$, where $y$ is the resource allocation as a function of valuation denoted by $v$. We observe that as the level of resource asymmetry increases, i.e., as $\alpha$ decreases from 1, the $\beta$ parameter of the exponential fit for the average best strategy against valuations illustrated in Figure \ref{fig10}, increases significantly. The disadvantaged player devotes an ever larger share of resources to the highest valued battlefield, and an ever decreasing share of resources over the lowest-valued battlefields. Such a concentration of resources is already a property of the equilibrium of the symmetric game illustrated by Figure \ref{fig10c}, dependent on the heterogeneity of valuations. What allows us to establish that asymmetric competition generates a stronger concentration of resources with respect to the symmetric benchmark, is that this concentration is larger than in the symmetric case. We establish that asymmetric competition in our genetic algorithm generates strategies with a higher concentration of resources. Towards which battlefields does this translates through? Figure panel \ref{fig10} suggests that concentration happens towards the highest-valued battlefields, as the disadvantaged player tries its best to at least minimize the score difference upon losing.

\subsection{Adaptation and counter strategies in heterogeneous Blotto games}
\label{results_sub4}

\begin{figure}[H]
     \centering
     \begin{subfigure}[b]{0.5\textwidth}
         \centering
         \includegraphics[width=\textwidth]{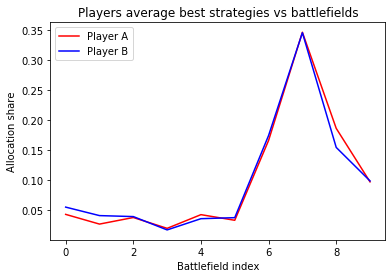}
         \caption{Against battlefield indexes}
         \label{fig11a}
     \end{subfigure}
     \hfill
     \begin{subfigure}[b]{0.5\textwidth}
         \centering
         \includegraphics[width=\textwidth]{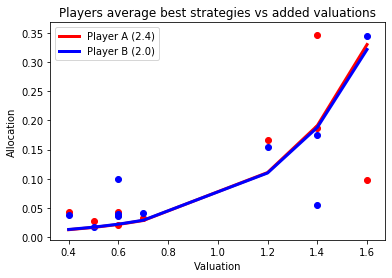}
         \caption{Against total valuations, exponential fit}
         \label{fig11b}
     \end{subfigure}
        \caption{Average best strategies in symmetric heterogeneous Blotto game after 1,000 iterations show correlation and adaptation of strategies}
        \label{fig11}
\end{figure}

Figure panel \ref{fig11} adds more insight to strategies developed under heterogeneous valuations and symmetric resources. Notably, players adapt to each other. For the first battlefield in the game, valued 0 by player A and 0.7 by player B, we would indeed expect player B in equilibrium to allocate more resources in this battlefield. However, if A does not value this territory at all, A should not invest much resources to conquer it. As a result, player B should decrease its allocation, to be able to compete in more valuable, or more contested battlefields elsewhere. We provide evidence that this mechanism exists in our algorithm. Figure \ref{fig11a} confirms visually that strategies are very correlated as a result. The Pearson correlation between the two average best strategies is over $0.992$. Competition seems to concentrate on very few battlefields. Numerical and graphical results illustrated by Figure \ref{fig11b} suggests that competition concentrates in battlefield of high total valuation (valuations of A added to valuations of B). The 8th battlefield of the game, valued respectively 0.7 and 0.9, appears the most desirable, and essential for score maximization, and for this reason, concentrates a major part of resources of both players.\\

\begin{figure}[H]
     \centering
     \begin{subfigure}[b]{0.5\textwidth}
         \centering
         \includegraphics[width=\textwidth]{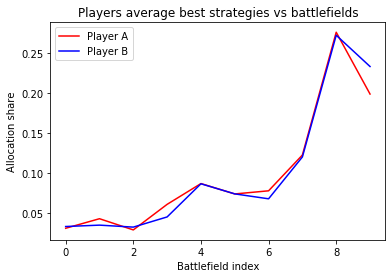}
         \caption{Against battlefield indexes}
         \label{fig11a}
     \end{subfigure}
     \hfill
     \begin{subfigure}[b]{0.5\textwidth}
         \centering
         \includegraphics[width=\textwidth]{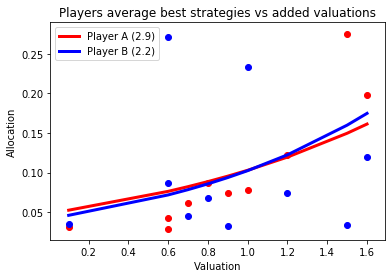}
         \caption{Against total valuations, exponential fit}
         \label{fig11b}
     \end{subfigure}
        \caption{Average best strategies in symmetric heterogeneous Blotto game after 1,000 iterations show correlation and adaptation of strategies}
        \label{fig11-2}
\end{figure}

We complement this point by showing under a different random seed, the similarity in the patterns of concentration of competition on a few specific battlefields, with a different draw of valuations vectors\footnote{$ [0. \ 0.1 \ 0.2 \ 0.3 \ 0.4 \ 0.5 \ 0.6 \ 0.7 \ 0.8 \ 0.9]$ and $[0.8 \ 0. \ 0.5 \ 0.3 \ 0.2 \ 0.7 \ 0.4 \ 0.9 \ 0.1 \ 0.6]$ respectively} to strengthen this point. At this effect, Figure panel \ref{fig11-2} shows concentration of competition on the last two battlefields, and a positive relation between total valuation and allocations. 

% [0.  0.1 0.2 0.3 0.4 0.5 0.6 0.7 0.8 0.9]
% [0.8 0.  0.5 0.3 0.2 0.7 0.4 0.9 0.1 0.6]

\begin{figure}[H]
     \centering
     \begin{subfigure}[b]{0.5\textwidth}
         \centering
         \includegraphics[width=\textwidth]{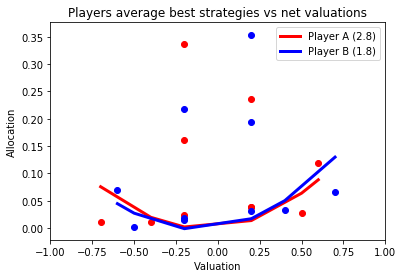}
         \caption{$\alpha = 1$}
         \label{fig12a}
     \end{subfigure}
     \hfill
     \begin{subfigure}[b]{0.5\textwidth}
         \centering
         \includegraphics[width=\textwidth]{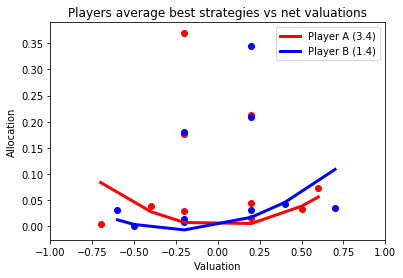}
         \caption{$\alpha = 0.9$}
         \label{fig12b}
     \end{subfigure}
        \caption{Average best strategies against net valuations after 1,000 iterations. Quadratic fit.}
        \label{fig12}
\end{figure}

In both symmetric and asymmetric Blotto games, heterogeneity of valuations generates \textit{counter strategies}. They consist for a player, in allocating resources on a battlefield that its opponent values more than it does, in order to prevent its opponent from winning it, denying the corresponding value from the opponent score. The quadratic fit provided in Figure panel \ref{fig12} suggests that such strategies do marginally exist in the heterogeneous Blotto game, either symmetric or asymmetric. \\

Interestingly, in a related fashion to how US Presidential Election campaign spending focuses on closely-valued, highly contested states such as Florida, these strategies heavily focus on similar battlefields. They display the feature that their valuations are very close for the two players, which translates into a low absolute net valuation. When valuations are very close for a given battlefield, competition seems to skyrocket, generating the same patterns as our stylized facts. Under this lens, we identify the seemingly irrational bidding behavior in US election where parties do not bid much in their highest-valued states, but sometimes a lot on middle-valued states. In our model, this behavior is consistent with players understanding each other's valuations, and reacting accordingly. Player A may think that player B values very highly a given battlefield, and that B is then ready to allocate a considerable force. As A does not value this battlefield as much, A will prefer to withdraw troops to allocate them in a more attractive, and less contested battlefield. As a result, B can progressively withdraw resources from this battlefield. This dynamic would generate the apparent zero bidding in the highest valued states. In our model, the differential between valuations acts similarly as would do a battlefield-specific probability of winning, biased towards a given player. Suppose that instead of the auction setting, victory is a continuous lottery function of allocations, and of a battlefield support probability for either player. Texas in this approach would be naturally more likely to give the victory to a Republican player. As a result, the Democrat player would probably withdraw any spending here, and focus instead Florida, which historically appears to propose equal victory chances to both candidates. \\

We could consider an extension of the model in this US election context, in which valuations for each state for each player are equal to $b_i^kv^k$, in which $b_i^k$ is the estimated player probability of victory in state $k$, and $v^k$ the number of Electoral College members in state $k$. In this way, we obtain heterogeneous valuations anchored in College members. Heterogeneity in this cases comes from heterogeneous assessments of the probability of winning, which depends of state fixed factors such as preferences, local electorate and political context for example. Empirically, this could be naively measured by looking at the frequency of victories in the past few elections. This will then result in the learning process, as in our results, in players ignoring some battlefields with a very high $v^k$, but a very low $b_i^k$: Texas for Democrats, California for Republicans. It will also, as in our simulations, and as in empirical patterns, generate high competition over battlefields with a lower $v^k$, but very close probabilities of victory $b_i^k$, such as Florida. We hope to further develop in some future work this extension of the model.

%We first observe the emergence of \textit{counter strategies} when the algorithms are trained to win rather than only maximizing their score. As they want to maximise the probability of having a higher score than their opponent, winning a battlefield that they value $x$ has the same impact as winning a battlefield their opponent values $x$. Under heterogeneous valuations, the genetic algorithm identifies quite efficiently which are the battlefields that have a high \textit{net valuation} for either of the players. That is, the battlefields for which there exists a large difference between the valuation given by player A, and the valuation given by player B. This gives rise, when plotting the outcome strategies against these players' net valuations, to U-shape distributions, as in Figure panel \ref{fig10}. The competition shifts from focusing your resources on the battlefields you value most, to focusing battlefields that have a high common value for the two players.

\section{Conclusion}
\label{section_conclusion}

In this article, we present and implement a genetic algorithm to study Blotto games with asymmetric resources between players, and heterogeneous valuations of battlefields that are common knowledge, a case that is not currently analytically solvable. We first establish that in the traditional Blotto game setting, in which players have the same resources, and similar valuations of the battlefields, our genetic algorithm identify optimal strategies that converge to the analytical Nash equilibrium. Running the algorithm on asymmetrical Blotto games with heterogeneous valuations, we highlight some interesting strategy insights, and their connections to stylized facts of empirical Blotto games. Notably, in situation of small resource asymmetry, the disadvantaged player tends to concentrate more its resources to compensate the difference. Under a large enough asymmetry of resources, the magnitude of the advantage sees the disadvantaged player shift its resources on the lowest valued battlefield to still minimize its loss. Heterogeneity of valuations generate more complex behavior, such as counter strategies and concentration of competition on battlefields with close valuations across players. The genetic algorithm is able to deliver both optimal strategies in these context beyond analytical study, and to generate some stylized facts of interests. By so doing, it sheds light on some strategic behavior in situations such as the US Presidential Election, seeing spending strategies in states as a Blotto game.\\

This work admittedly does not constitute an exhaustive study of Blotto games. In particular, results operate under the assumption that players' valuations for battlefields, albeit heterogeneous, are considered to be fixed. In game theoretical terms, they are known by the agents, and constitute common knowledge. In the algorithm, players quickly understand their valuations, and the ones of their opponents. In several empirical situations adequately modeled by Blotto games, valuations are not so fixed and obvious. Studying Blotto games under imperfect, or even absence of information, may contribute to a grater understanding of these empirical; strategical dilemmas.\\

Finally, we outline that the method of genetic algorithm used here, albeit not completely new to game theory, is novel in its application to the problem of Blotto Games. The ability of the genetic algorithm to, under a very finite time, and computationally capacity, efficiently scan very large space of possible strategies and to identify strategies that are as good as, and even contextually more performant, than the analytical Nash equilibrium, can be used to solve other strategic problems of great complexity. We think notably of an operational application to empirical Blotto games, such as the US Presidential elections, but also to other games such as Attack Defence games on networks, that can be considered as an extension of Blotto games in which battlefields and resources are connected by a network structure. In this more complex situation closer to cybersecurity and war-games, the genetic algorithm may contribute to obtain more performant attack and defence strategies, as well as more robust network architecture for security, or even for more resilient economic and financial systems.

\vspace*{\fill}

\section*{Compliance with Ethical Standards}
\textbf{Funding}: the author has no funding to report.\\
\textbf{Conflict of Interest}: The author declare that he has no conflict of interest.

\section*{Acknowledgments}
The author wishes to warmly thank Prof. Compte for his help and advice throughout this project, as well as the participants and organizers of the Economic Theory research seminar at the Paris School of Economics, for their useful remarks and suggestions. 

\clearpage
\section{References}
\begingroup
\renewcommand{\section}[2]{}%

\endgroup
\clearpage

\clearpage
\end{document}